# Dempster-Shafer vs. Probabilistic Logic

Daniel Hunter
Northrop Research and Technology Center
One Research Park
Palos Verdes Peninsula CA. 90274

**Abstract**

The combination of evidence in Dempster-Shafer theory is compared with the combination of evidence in probabilistic logic. Sufficient conditions are stated for these two methods to agree. It is then shown that these conditions are minimal in the sense that disagreement can occur when any one of them is removed. An example is given in which the traditional assumption of conditional independence of evidence on hypotheses holds and a uniform prior is assumed, but probabilistic logic and Dempster's rule give radically different results for the combination of two evidence events.

## 1 Introduction

Researchers on uncertain reasoning within the AI community have recently shown interest in probabilistic reasoning using sets of standard probability assignments. For example, Nilsson in [6] and Grosof in [3] have considered methods for reasoning with sets of probability assignments generated by probabilistic equality and inequality constraints[1]. Following Nilsson, I use the expression "Probabilistic Logic" to denote the collection of such methods. The aim of these methods is to compute a set of possible probabilities for a given statement from the specified set of probability assignments. If the set of probability assignments is generated by probabilistic equality and inequality constraints, the possible probabilities for a given statement form an interval. Since Dempster-Shafer also associates an interval with each statement A, namely the interval bounded by Bel(A) and Pls(A), the question arises as to the connection between Dempster-Shafer belief functions and sets of probability assignments defined by equality and inequality constraints. Grosof [3] has shown that the latter is a generalization of the former: every Dempster-Shafer belief function is representable by a set of probability assignments arising from equality and inequality constraints, but not vice-versa. A related issue concerns the connection between Dempster's rule of combination and the combination of evidence statements in probabilistic logic. Grosof [2] states some results concerning conditions under which these two methods of combining evidence yield the same result. The aim of this paper is to generalize Grosof's results and to investigate how divergent the two

---
[1] I.J. Good has been an advocate of probabilistic reasoning using inequality statements since about 1938. See [1, p.25 and pp.75-76]

methods can become when the conditions for agreement are not satisfied. Familiarity with the basics of Dempster-Shafer theory is assumed.

## 2 Conditions for Agreement

Recall that where $m_1$ and $m_2$ are two mass functions, with focal elements $A_1, ..., A_k$ and $B_1, ..., B_l$, respectively, their combination $m_1 \bigoplus m_2$, called their *orthogonal sum*, is defined by:

$$m_1 \bigoplus m_2(A) = \frac{\sum_{A_i \cap B_j = A} m_1(A_i) m_2(B_j)}{1 - \sum_{A_i \cap B_j = \emptyset} m_1(A_i) m_2(B_j)}$$

$$= \frac{\sum_{A_i \cap B_j = A} m_1(A_i) m_2(B_j)}{\sum_{A_i \cap B_j \neq \emptyset} m_1(A_i) m_2(B_j)}$$

And where m is a mass function, the belief function Bel determined by m is given by:

$$Bel(A) = \sum_{B \subseteq A} m(B)$$

When working within probabilistic logic, I follow Grosof in making explicit the evidence on which a mass function depends. Hence for each mass function $m_i$, the statement $m_i(A) = p$ within Dempster-Shafer has as counterpart in probabilistic logic the statement $P(A|E_i) = p$, where P is a standard probability function and $E_i$ is a statement representing the evidence on which $m_i$ is based.

The following is the most general theorem I know of that states conditions under which application of Dempster's rule agrees with combination of evidence in probabilistic logic:

**Theorem 1** *Let $m_1$, $m_2$ be mass functions over frame $\Theta$ each with focal elements $S_1, ..., S_k$, where the $S_i$ form a partition of $\Theta$ (i.e., $S_i \cap S_j = \emptyset$ for $i \neq j$ and $S_1 \cup ... \cup S_k = \Theta$); $E_1$ and $E_2$ propositions not defined in $\Theta$; $\mathcal{E} = \{E_1 \wedge E_2, E_1 \wedge \neg E_2, \neg E_1 \wedge \neg E_2\}$; and $\Gamma$ a set of probability assignments P over $\Theta \times \mathcal{E}$ satisfying*

(i) $P(S_i) = 1/k$, $i = 1,...,k$. *(By abuse of notation, I identify $X \subseteq \Theta$ with $X \times \mathcal{E} \subseteq \Theta \times \mathcal{E}$.*

(ii) $P(E_1 \wedge E_2 | S_i) = P(E_1|Si) P(E_2|Si)$, $i = 1,...,k$.

(iii) $P(S_i|E_1) = m_1(S_i)$ *and* $P(S_i|E_2) = m_2(S_i)$, $i = 1,...,k$.

(iv) $P(E_1 \wedge E_2) > 0$.

*Then, where $Bel_{1,2}$ is the belief function over $\Theta$ determined by $m_3 = m_1 \bigoplus m_2$, for all $A \subseteq \Theta$, $i \in \{1, ..., k\}$, and $R \in \Gamma$:*

$$Bel_{1,2}(S_i) = m_3(Si) = R(S_i | E_1 \wedge E_2) \tag{1}$$

*and*

$$Bel_{1,2}(A) = min\{Q(A|E_1 \wedge E_2) : Q \in \Gamma\} \tag{2}$$

(Grosof [2] states the first part of theorem 1 for a two-membered partition; both parts of the theorem can also be derived from results in Yen [7][2] .)

**Proof.** First we prove equation (1): For each $S_j$ and $P \in \Gamma$, we have

$$\begin{aligned} P(S_j|E_1 \wedge E_2) &= \frac{P(E_1 \wedge E_2|S_j)P(S_j)}{\sum_{i=1}^{k} P(E_1 \wedge E_2|S_i)P(S_i)} = \frac{P(E_1|S_j)P(E_2|S_j)}{\sum_{i=1}^{k} P(E_1 \wedge E_2|S_i)} \\ &= \frac{[P(E_1)P(S_j|E_1)/P(S_j)][P(E_2)P(S_j|E_2)/P(S_j)]}{\sum_{i=1}^{k}[P(E_1)P(S_i|E_1)/P(S_i)][P(E_2)P(S_i|E_2)/P(S_i)]} \\ &= \frac{P(S_j|E_1)P(S_j|E_2)}{\sum_{i=1}^{k} P(S_i|E_1)P(S_i|E_2)} = \frac{m_1(S_j)m_2(S_j)}{\sum_{i=1}^{k} m_1(S_i)m_2(S_i)} \\ &= m_3(S_j) = Bel_{1,2}(S_j). \end{aligned}$$

To prove equation (2) we construct a $P \in \Gamma$ such that $P(A|E_1 \wedge E_2) = min\{Q(A|E_1 \wedge E_2 : Q \in \Gamma\}$ and $P(A|E_1 \wedge E_2) = Bel_{1,2}(A)$. To do so we must distinguish between $X \subseteq \Theta$ and $X \times \mathcal{E} \subseteq \Theta \times \mathcal{E}$. Let $A \subseteq \Theta$. We wish to construct a probability function P over $\Theta \times \mathcal{E}$ such that $P(A \times \mathcal{E}|E_1 \wedge E_2) = min\{Q(A \times \mathcal{E}|E_1 \wedge E_2) : Q \in \Gamma\}$. The desired probability function P will be determined if $P(\theta \wedge e)$ is defined for each $\theta \in \Theta, e \in \mathcal{E}$. This will be accomplished if for each $S_i$, P is defined for every element of $S_i \times \mathcal{E}$. Pick any R from $\Gamma$ (if $\Gamma$ is empty, the theorem is vacuously true). For each $S_i$, define P over the elements of $S_i \times \mathcal{E}$ as follows: if $S_i \subseteq A$, set $P(\theta \wedge e) = R(\theta \wedge e)$ for each $\theta \wedge e \in S_i \times \mathcal{E}$; otherwise, choose a $\theta_0 \in S_i - A$ and set $P(\theta_0 \wedge e) = R(S_i \wedge e)$ and $P(\theta \wedge e) = 0$ for all $\theta \in S_i, \theta \neq \theta_0$. This fixes P for all singletons in $\Theta \times \mathcal{E}$. By the construction of P, we have $P(S_i \wedge e) = R(S_i \wedge e)$ and therefore since R satisfies (i)-(iv), so does P. Hence $P \in \Gamma$. It is easy to verify that:

$$P(A \times \mathcal{E}|S_i \wedge X) = \begin{cases} 1 & \text{if } S_i \subseteq A \\ 0 & \text{otherwise} \end{cases}$$

We now wish to show that $P(A \times \mathcal{E}|E_1 \wedge E_2) = min\{Q(A \times \mathcal{E}|E_1 \wedge E_2) : Q \in \Gamma\} = Bel_{1,2}(A)$. By probability theory, for any probability assignment Q in $\Gamma$, $Q(A \times \mathcal{E}|E_1 \wedge E_2) = \sum_{i=1}^{k} Q(S_i|E_1 \wedge E_2)Q(A \times \mathcal{E}|S_i \wedge E_1 \wedge E_2)$. By equation (1), $Q(A \times \mathcal{E}|E_1 \wedge E_2) = \sum_{i=1}^{k} m_3(S_i)Q(A \times \mathcal{E}|S_i \wedge E_1 \wedge E_2)$. If $S_i \subseteq A$, then $Q(A \times \mathcal{E}|S_i \wedge E_1 \wedge E_2) = 1$. Hence $Q(A \times \mathcal{E}|E_1, E_2)$ will be minimal if $Q(A \times \mathcal{E}|S_i \wedge E_1 \wedge E_2) = 0$ when $S_i$ is not a subset of A. But P has this property, so

$$min\{Q(A \times \mathcal{E}|E_1 \wedge E_2) : Q \in \Gamma\} = P(A \times \mathcal{E}|E_1 \wedge E_2) =$$
$$\sum_{S_i \subseteq A} P(S_i \times \mathcal{E}|E_1 \wedge E_2) = \sum_{S_i \subseteq A} m_3(S_i) = Bel_{1,2}(A). \square$$

To avoid misunderstanding, I should emphasize that the above theorem only states *sufficient*, not necessary, conditions for use of Dempster's rule to agree with combination of evidence in probabilistic logic. Thus it is quite possible for there to be cases in which the two methods of combination agree, but not all, and possibly none, of the above sufficient conditions for agreement are satisfied. However, three points need to be made here: first, as far as I know, no non-trivial *necessary* conditions for agreement have yet been stated (not even condition (ii), the independence condition, is necessary); second, if we think that probabilistic logic gives the right answer but wish to use Dempster's rule for computational convenience, then in order to be sure that a particular application of Dempster's rule gives the right answer, we need sufficient conditions for agreement, since the satisfaction of merely necessary conditions for agreement is no guarantee that there is agreement. Finally, what is in effect shown below is that the conditions of theorem 1 form a *minimal* set of sufficient conditions in the sense that if any one of them is removed then the theorem no longer holds.

---

[2]Yen in [7] is not directly concerned with probabilistic logic; however, his theorem 1 can be interpreted as applying to a class of probability functions and by adding the equivalent of my assumption (i) to Yen's assumptions, it is not hard to show that theorem 1 of the present paper follows.

## 3  How Much Disagreement?

The next question that arises is, How much divergence arises between Dempster-Shafer and probabilistic logic if one or more of the conditions of the theorem is not satisfied? Obviously condition (iii) on P must be kept and (iv) is necessary for the conditional probabilities to be defined. Thus the obvious candidates for scrutiny are conditions (i) and (ii). But other, less obvious, assumptions also enter into the theorem: for example, it is assumed that the focal elements of $m_1$ and $m_2$ are the same and that they constitute a partition of $\Theta$. This section shows that lifting any one of these assumptions can result in dramatic disagreement between Dempster-Shafer and probabilistic logic.

Let us begin by examining the effect of lifting the assumption that the members of the partition are equally probable. If (i) were abandoned, then the prior over the $S_i$ could swamp the effect of $E_1$ and $E_2$. For example, given any fixed values for $P(S_i|E_1)$ and $P(S_i|E_2)$, providing both these values are strictly between zero and one, $P(S_i|E_1 \wedge E_2)$ can take on any value strictly between zero and one depending upon the value of $P(S_i)$. For simplicity consider the case of a bipartite partition of $\Theta$ – i.e. there are only two members to the partition, call them $H$ and $\overline{H}$. Then if conditions (ii) and (iv) hold for P, the formula

$$P(H|E_1 \wedge E_2) = \frac{P(H|E_1)P(H|E_2)}{P(H|E_1)P(H|E_2) + O(H)P(\overline{H}|E_1)P(\overline{H}|E_2)}$$

can be proven. The factor O(H) in the second term of the denominator is the *odds on H*, defined to be $P(H)/P(\overline{H})$. By making O(H) sufficiently high, the denominator can be made large, thus bringing $P(H|E_1 \wedge E_2)$ close to zero, regardless of the values of $P(H|E_1$ and $P(H|E_2)$ (providing neither is equal to one). However, if the sum of $m_1(H)$ and $m_2(H)$ is greater than one, $m_1 \oplus m_2(H)$ will be greater than either $m_1(H)$ or $m_2(H)$. For example, with $m_1(H) = m_2(H) = P(H|E_1) = P(H|E_2) = 0.9$, but with $P(H) = 0.999$, we get $m_1 \oplus m_2(H) \approx 0.99$ but $P(H|E_1 \wedge E_2) = 0.075$, a rather large difference indeed. Similarly, making $O(H)$ small results in $P(H|E_1 \wedge E_2)$ being close to one.

The above example presents a counter-intuitive consequence of standard probability theory: the higher the prior probability of a hypothesis, the lower will be its posterior probability on the basis of the conjunction of two evidence statements that are conditionally independent under both the hypothesis and its negation. Though counterintuitive, this consequence can be made more plausible by considering the ratio $P(H|E)/P(H)$, of the posterior probability to the prior, and noting that the higher the prior, the smaller this ratio and so the less confirmatory the evidence is of the hypothesis. In particular, if P(H) is higher than $P(H|E)$, then even if $P(H|E)$ is high, E will be evidence *against* H and so the effect of combining two such evidence statements, when they are conditionally independent, is to even further lower the posterior probability of H.

Dempster's rule also diverges from probabilistic logic when the evidence statements are not conditionally independent under the members of the partition. It is well known that the combined effect of two non-independent evidence statements is not determined by their individual effects on the probability of a hypothesis (except when one of the posterior probabilities is zero or one). I will therefore say no more about the consequences of lifting condition (ii).

Of more interest is the question of what happens when the conditions on P are maintained, but the conditions for the mass function are changed. Recall that it was assumed that both mass functions have the same focal elements and that these focal elements form a partition of the frame of discernment. Consider the latter condition first. What if the focal elements do not form a partition? In this case, the main difference is that under Dempster's rule, intersections of focal elements always obtain some mass in the combined mass distributions, but the same intersections do not always have a positive probability in the posterior on the basis of the both evidence statements. For example, suppose that $\Theta = \{a, b, c\}$ and $m_1\{a,b\} = m_2\{a,b\} = m_1\{b,c\} = m_2\{b,c\} = 0.5$. Then $m_3\{b\} = 0.5$, but it is easy to construct a P satisfying (i)-(iv) such that $P(b|E_1 \wedge E_2) = 0$ – e.g., set $P(b) = 0$ and $P(a) = P(c) = P(E_1) = P(E_2) = P(a|E_1) = P(c|E_2) = 0.5$.

I will present one final example in which Dempster's rule diverges from probabilistic logic, one that in my opinion shows a serious defect in Dempster's rule. In this example, the focal elements for each mass function form a partition but not the same partition. Let the frame of discernment $\Theta = \{x_1, ..., x_n\}$ and let the focal elements of $m_1$ be $\{x_1\}$ and $\{x_2, ..., x_n\}$ and the focal elements of $m_2$ the singleton elements of $\Theta$. Assume $m_2(\{x_i\}) = 1/n, i = 1, ..., n$. Then

$$m_3(\{x_1\}) = \frac{m_1(\{x_1\})m_2(\{x_1\})}{m_1(\{x_1\})m_2(\{x_1\}) + \sum_{i=2}^{n} m_1(\{x_2,...,x_n\})m_2(\{x_i\})} \quad (3)$$

$$= \frac{m_1(\{x_1\})1/n}{m_1(\{x_1\})1/n + (1 - m_1\{x_1\})\sum_{i=2}^{n} 1/n} \quad (4)$$

$$= \frac{m_1(\{x_1\})}{m_1(\{x_1\}) + (1 - m_1(\{x_1\}))\sum_{i=2}^{n} 1} \quad (5)$$

$$= \frac{m_1(\{x_1\})}{m_1(\{x_1\}) + (1 - m_1(\{x_1\}))(n - 1)} \quad (6)$$

It can be seen that $m_3(\{x_1\})$ goes to zero as n goes to infinity, providing $m_1(\{x_1\}) < 1$. This is a disconcerting result. To see why, consider a concrete case in which the above mass functions might be combined. Suppose there is a lottery with n individuals participating and only one winner. Let the frame of discernment be $\{x_1, ..., x_n\}$, where $x_i$ is the event of the $i^{th}$ participant (in some ordering of the participants) winning. It is known beforehand what the winning number is. One piece of evidence is that Jones holds a ticket whose digits are identical with those of the winning number, except possibly for one digit (e.g, you see Jones' ticket except for one digit, which is obscured). Another piece of evidence is that the lottery is fair: the participants get their tickets

26

through some random drawing process. In the Dempster-Shafer theory, the first piece of evidence, in the absence of the second, would plausibly be represented by a mass distribution of the form of $m_1$ – e.g. if $x_1$ is the event of Jones' winning the lottery, then we might set $m_1(\{x_1\}) = 0.1$ if we see that Jones' ticket is identical with the winning ticket except possibly for one digit and, in the absence of knowledge as to whether or not the lottery is fair, Dempster-Shafer would presumably recommend spreading the remaining mass over the set $\{x_2, ..., x_n\}$, without assigning any mass to smaller subsets. And the second piece of evidence, in the absence of the first, would, I should think, be represented by $m_2$ since we have positive evidence that each participant has an equal chance of winning.

With n = 112 and $m_1(x_1) = 0.1$, we have

$$m_3(\{x_1\}) = 0.001,$$

which, being the total mass committed to $\{x_1\}$, yields

$$Bel_{1,2}(\{x_1\}) = 0.001.$$

But this degree of belief seems much too low: if you believe that Jones' has at least a 1 in 10 chance of winning the lottery on the basis of seeing all but one digit of Jones' ticket, learning that the lottery is fair should not cause you to lower your degree of belief in Jones' winning. Worse still, since combination of evidence is commutative in Dempster-Shafer, imagine first learning that the lottery is fair, in which case you assign a 1 in 112 chance that Jones will win, and then learning that all but possibly one of the digits in Jones' ticket match those in the winning number. Surely it would be absurd to then lower Jones' chances of winning to 1 in 1000.

How would probabilistic logic handle the same example? Note that we cannot really keep the conditions on the probability assignment P in theorem 1 the same, since they refer to the $S_i$, which are stipulated to be focal elements for both $m_1$ and $m_2$. However, we can assume that $P(F|E_1) = m_1(F)$ for each focal element of $m_1$ and similarly for $m_2$. Also, condition (i) presents a bit of a problem since for k > 2, (i) cannot apply to both sets of focal elements. We assume instead that (i) applies to the singletons of $\Theta$. In short, we assume that P satisfies the following conditions:

(1) $P(\{x_i\}) = 1/n, i = 1, ..., n.$

(2a) $P(E_1 \wedge E_2|\{x_i\}) = P(E_1|\{x_i\})P(E_2|\{x_i\}), i = 1, ..., n.$
(2b) $P(E_1 \wedge E_2|\{x_2, ..., x_n\}) = P(E_1|\{x_2, ..., x_n\})P(E_2|\{x_2, ..., x_n\})$

(3a) $P(\{x_1\}|E_1) = m_1(\{x_1\})$
(3b) $P(\{x_i\}|E_2) = 1/n, i = 1, ..., n.$

(4) $P(E_1 \wedge E_2) > 0.$

Conditions (1)-(4) entail:

$$P(\{x_i\}|E_1 \wedge E_2) = \frac{P(\{x_1\}|E_1)P(\{x_1\}|E_2)}{\sum_{i=1}^{n} P(\{x_i\}|E_1)P(\{x_i\}|E_2)} = P(\{x_1\}|E_1). \qquad (7)$$

27

Hence $E_2$ has no effect on the probability of $\{x_1\}$ in the presence of $E_1$ (in fact, as required by the theorem of Johnson [4, p.199] for the case $n > 2$, $E_2$ is irrelevant to any $x_i$). Hence no matter how large n is, the probability of Jones' winning given both $E_1$ and $E_2$ will be 0.1. This seems a much more reasonable result.

An objection to the above comparison of Dempster-Shafer with probabilistic logic was raised by one of the reviewers of this paper. According to this objection, there is nothing surprising in the fact that the combination of the evidence about Jones' ticket and the evidence about the fairness of the lottery lowers Jones' probability of winning. After all, both pieces of evidence state that it is highly unlikely that Jones will win, so why shouldn't their combination make it even more unlikely that he will win?

This objection confuses a hypothesis's being unlikely on the basis of certain evidence with its being disconfirmed by that evidence. A piece of evidence *disconfirms* a given hypothesis if the probability of that hypothesis on the basis of that piece of evidence is lower than the prior probability of the hypothesis. If two independent pieces of information disconfirm a hypothesis, then their conjunction should indeed disconfirm the same hypothesis to an even greater degree. However, in the above example, the evidence about Jones' ticket does not disconfirm the hypothesis that he will win. To the contrary, given the assumed size of the hypothesis space, it significantly increases Jones' probability of winning. Furthermore, the example can modified so that the evidence about Jones' ticket makes it highly probable that he will win: assume that you see all the digits in Jones' ticket and are ninety percent certain that Jones holds the winning ticket (you may be slightly unsure about one of the digits in the winning ticket). Then if the only modification to the example is that $m_1(\{x_1\}) = 0.9$, we find that $Bel_{1,2}(\{x_1\})$ is 0.075, still much too low a number, while $P(\{x_1\}|E_1 \wedge E_2) = 0.9$.

The source of the discrepancy between Dempster's rule and probabilistic logic in this case can be discovered by rewriting the equations for $m_3(\{x_1\})$ and $P(\{x_1\}|E_1 \wedge E_2)$ as follows:

$$m_3(\{x_1\}) = \frac{m_1(\{x_1\})}{m_1(\{x_1\}) + T_1} \tag{8}$$

$$T_1 = \sum_{i=2}^{n} m_1(\{x_2,...,x_n\}) \tag{9}$$

$$P(\{x_1\}|E_1 \wedge E_2) = \frac{P(\{x_1\}|E_1)}{P(\{x_1\}|E_1) + T_2} \tag{10}$$

$$T_2 = \sum_{i=2}^{n} P(\{x_i\}|E_1) \tag{11}$$

The difference is in the terms $T_1$ and $T_2$. The difference is that $m_1(\{x_2,...,x_n\}$ is a constant, whereas $P(\{x_i\}|E_1), i = 2,...,n$ grows, *on average*, smaller as n increases since the term $T_2$ is equal to $P(\{x_2 ..., x_n\}|E_1)$, which is stipulated to be equal to $m_2(\{x_2,...,x_n\})$, a constant. Hence $T_1$ goes to infinity as n goes to infinity, but $T_2$ remains constant.

## 4 Conclusion

I have proven that Dempster's rule of combination agrees with combination of evidence in probabilistic logic under certain conditions. I have also shown that these two methods for combining evidence can produce radically different results when these conditions do not obtain. Of particular interest is the fact that even when the conditional independence assumptions are satisfied, differences can result when the focal elements of the two mass functions do not form a partition or form different partitions.